\documentclass{article}
\usepackage{iclr2027_conference,times}

\usepackage[utf8]{inputenc} 
\usepackage[T1]{fontenc}    
\usepackage{hyperref}       
\usepackage{url}            
\usepackage{booktabs}       
\usepackage{amsfonts}       
\usepackage{nicefrac}       
\usepackage{microtype}      
\usepackage{xcolor}         

\usepackage[ruled, vlined, linesnumbered]{algorithm2e}
\usepackage{amsmath,amsthm,amssymb}
\usepackage{bbm}
\usepackage{bm}
\usepackage[capitalize]{cleveref}
\usepackage{enumitem}
\usepackage{graphicx}
\usepackage{lipsum}  
\usepackage{multirow}
\usepackage{makecell}
\usepackage{pgfplots}
\usepackage{placeins}
\usepackage{subcaption}
\usepackage{soul}
\usepackage{tabularx}
\usepackage{wrapfig}
\usepackage{xspace}

\usepackage{listings}
\usepgfplotslibrary{groupplots,dateplot}
\usetikzlibrary{patterns,shapes.arrows}
\pgfplotsset{compat=newest}

\usepackage{tcolorbox}
\newtcolorbox{prompt}[1]{colback=gray!20,colframe=gray!50!black,fonttitle=\bfseries,title=#1}

\newcommand{\ours}{SIPO}
\newcommand{\markblue}[1]{\textcolor{black}{#1}}

\title{\ours{}: Unifying Reinforcement Learning with On-Policy Self-Distillation}

\author{Zhenrui Yue$^1$, Huimin Zeng$^1$, Yueqi Wang$^2$, Yaokun Liu$^1$, Fengran Mo$^3$, Jinghan Zhang$^4$, \\
\textbf{Mung Yao Jia$^1$, Gyuseok Lee$^1$, Yang Zhang$^5$, Na Wei$^1$, Dong Wang$^1$} \\
$^1$University of Illinois, $^2$UC San Diego, $^3$Rochester Institute of Technology \\$^4$Clemson University, $^5$Miami University \\
\texttt{\{zhenrui3, dwang24\}@illinois.edu} \\
}

\iclrfinalcopy 

\begin{document}

\maketitle
\lhead{Preprint.}

\begin{abstract}
\markblue{
Reinforcement learning with verifiable rewards (RLVR) has become a standard paradigm for improving large language models (LLMs) on various tasks, yet its sparse outcome rewards lack token-level credit assignment for intermediate steps. To address this, on-policy self-distillation (OPSD) leverages a self-teacher with privileged context to provide additional dense learning signals. However, because the self-teacher is often overconfident and imposes excessive penalties on long reasoning trajectories, OPSD frequently struggles in practice. To mitigate this, we propose self-instructing policy optimization (\ours{}) with a contrastive self-teacher to provide dense credit. At each iteration, \ours{} samples multiple rollouts per prompt from the current policy, scores them with environment rewards, and constructs two teacher contexts for each rollout by pairing the reference answer with mistakes made within the group. The model then re-evaluates its own responses under both contexts, using the difference between the two teacher log-probabilities as token-level feedback, so that biases shared by both contexts are expected to largely cancel. The resulting objective yields a token-level advantage for every rollout: the reward still sets the main direction of each update while the self-teacher redistributes credit across tokens. Even in groups where every rollout fails and group-relative advantages vanish, \ours{} still provides a learning signal. By preserving direct optimization of the task reward while providing dense, token-level feedback, this approach bridges reinforcement learning and on-policy self-distillation. Extensive experiments across multiple reasoning and code-generation benchmarks demonstrate that \ours{} outperforms both RLVR and OPSD baselines without an external teacher or additional generation.
}
\end{abstract}

\section{Introduction}

Large language models (LLMs) have demonstrated strong capabilities in complex reasoning tasks~\citep{yang2025qwen3, comanici2025gemini, singh2025openai}. A critical mechanism behind these recent advancements is reinforcement learning with verifiable rewards (RLVR), which fine-tunes LLMs using outcome-level feedback~\citep{lambert2024tulu, guo2025deepseek,zhang2026starpo}. By directly optimizing for task rewards, RLVR enables models to discover reasoning strategies beyond those present in supervised data. Among RLVR methods, group relative policy optimization (GRPO) has emerged as a widely adopted approach that eliminates the need for a learned critic by estimating advantages from multiple rollouts per prompt~\citep{shao2024deepseekmath}.

However, a fundamental limitation of the RLVR paradigm remains: outcome rewards are inherently sparse. A single scalar reward assigned to an entire trajectory fails to provide token-level credit assignment, making it difficult for the model to identify which reasoning steps contributed to success or failure~\citep{lightman2023let, zhang2025lessons}. This sparsity is most severe on challenging problems: when every sampled rollout fails, group-relative advantages vanish and the prompt provides no learning signal at all, leaving the compute spent on sampling these rollouts unused. When training Qwen3-8B on DAPO-Math-17k, such uniformly failed groups account for 20--30\% of all groups early in training and still about 10\% at the end (see \Cref{sec:exp}).

\begin{figure}[t]
    \centering
    \includegraphics[trim=4.2cm 4.4cm 4.2cm 4.4cm, clip, width=\textwidth]{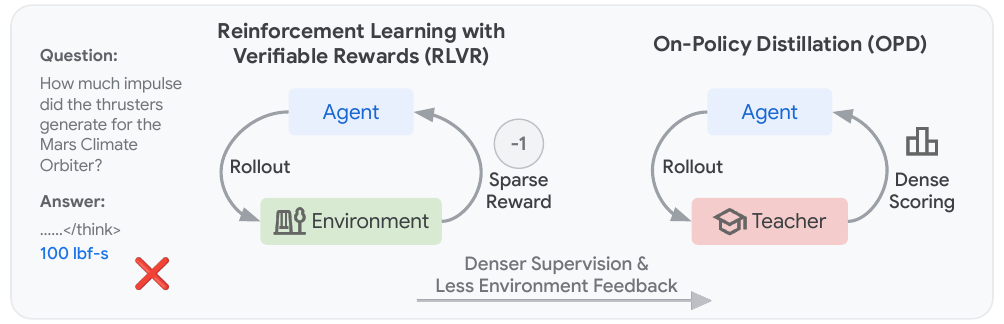}
    \caption{\textbf{Reinforcement learning versus on-policy distillation.} Reinforcement learning learns from sparse, task-grounded environment rewards, whereas on-policy distillation scores the student's own rollouts with a teacher, providing denser supervision but less direct environment feedback.}
    \label{fig:intro}
\end{figure}

Recent work on on-policy distillation (OPD) addresses this gap by providing dense, token-level supervision with an external teacher~\citep{agarwal2024policy, lu2025onpolicydistillation}. As illustrated in \Cref{fig:intro}, the student policy learns from its own rollouts: a more capable teacher model evaluates these trajectories to produce a token-level target distribution, which the student is then trained to match. However, OPD requires an external teacher to score the student's rollouts and minimizes a divergence objective rather than directly optimizing for task rewards; consequently, it underperforms without a strong teacher model, as the student's learning potential is upper-bounded by the capabilities of the teacher~\citep{li2026rethinking}. On-policy self-distillation (OPSD)~\citep{zhao2026self} instead conditions the model itself on privileged information to act as the teacher, but such a self-teacher is often overconfident and imposes excessive penalties on long reasoning trajectories (see \Cref{fig:sdpo_math}). Moreover, in scenarios where successful rollouts are rarely sampled or the teacher fails to provide meaningful supervision, the distillation signal becomes sparse or even absent~\citep{hubotter2026reinforcement}.

To bridge this gap, we introduce self-instructing policy optimization (\ours), a simple and effective RL framework that unifies reward-based policy gradients and on-policy self-distillation through a single token-level advantage. Specifically, at each iteration, \ours{} samples multiple rollouts from the current policy, scores them with the reward function, and constructs two teacher contexts for each rollout: the reference answer and a mistake from the group, such as the rollout's own wrong answer. Following OPSD, the model re-evaluates its own responses under both contexts without generating new text. To counter the self-teacher's overconfidence, we construct contrastive self-teachers~\citep{heakl2026cepo, pan2026rlcsd} and use the difference between the two teacher log-probabilities as token-level feedback, so that biases shared by both contexts are expected to largely cancel. A token thus gains credit when it is more likely under the correct context than under the mistaken one. The resulting objective then adds this clipped contrastive feedback to the group-relative advantage, yielding a token-level advantage for every rollout: the reward still sets the main direction of each update while the self-teacher redistributes credit across tokens. Importantly, since the contrast needs no successful rollout, uniformly failed groups, which lack any RL signal and are left untouched by prior contrastive methods, still contribute to learning via the self-distillation feedback, thereby improving sample efficiency.
We validate \ours{} across multiple reasoning and code-generation benchmarks, where it outperforms both RLVR and OPSD baselines without requiring an external teacher model or additional generation at training time. We summarize our contributions as follows:\footnote{Our code is provided at https://github.com/Yueeeeeeee/SIPO.}
\begin{itemize}[leftmargin=20pt]
    \item We propose self-instructing policy optimization (\ours), which unifies reward-based policy gradients and on-policy self-distillation through a single token-level advantage. Specifically, \ours{} adds contrastive self-teacher feedback to the advantage of every rollout, so that even groups in which every rollout fails still provide a learning signal.
    \item For the contrastive self-teacher, we construct two teacher contexts by pairing the reference answer with mistakes made within each prompt group, incorporating environment feedback when available. This enables the model to derive dense, token-level signals directly from its on-policy generations, eliminating the need for an external teacher.
    \item Extensive experiments on STEM reasoning and code-generation benchmarks demonstrate that \ours{} outperforms both RLVR and OPSD baselines under comparable computational budgets. Further analyses suggest that the contrastive self-teacher mitigates the overconfidence of a standard self-teacher, and provides a useful learning signal to uniformly failed groups.
\end{itemize}
\section{Related Work}

\subsection{Reinforcement Learning}
Reinforcement learning (RL) is a paradigm where an agent interacts with an environment and learns to make decisions that maximize cumulative rewards~\citep{sutton1998reinforcement}. Recently, RL has been adopted to improve LLMs through reinforcement learning from human feedback (RLHF)~\citep{ouyang2022training}. Such fine-tuning typically employs policy gradient algorithms like REINFORCE~\citep{sutton1999policy}. To reduce variance, actor-critic methods like A2C~\citep{mnih2016asynchronous} leverage learned critic networks for advantage estimation. Building on this, proximal policy optimization (PPO)~\citep{schulman2017proximal} bounds policy updates via a clipped surrogate objective, achieving improved training stability and robustness. Alongside these approaches, direct preference optimization (DPO)~\citep{rafailov2023direct} aligns language models using pairwise human preference comparisons. More recently, reinforce leave-one-out (RLOO)~\citep{ahmadian2024back} proposes to generate multiple responses and uses the mean reward of the remaining responses as a baseline. Similarly, group relative policy optimization (GRPO)~\citep{shao2024deepseekmath} and its variants like REINFORCE++ and DAPO~\citep{hu2025reinforce++, yu2025dapo} compute baselines from group-level or batch-level reward statistics across candidate completions, reducing memory overhead while maintaining stability. However, a fundamental limitation of these RLVR methods is that outcome rewards assign a single scalar to an entire trajectory, providing coarse credit assignment at the token level. In this work, we address this limitation by unifying reward-based policy gradients with on-policy distillation, enabling dense, token-level supervision while preserving direct optimization using the task reward.

\subsection{On-Policy Distillation}
On-policy distillation (OPD) addresses RLVR's sparse credit assignment by providing dense, token-level learning signals~\citep{agarwal2024policy, lu2025onpolicydistillation}. In OPD, a capable teacher evaluates the student's on-policy rollouts to produce token-level target distributions; the student is then trained to minimize its divergence from these targets. \citet{agarwal2024policy} introduce a generalized framework for OPD that studies various divergence measures and demonstrates that on-policy training can mitigate distribution mismatches in offline distillation. More recently, \citet{lu2025onpolicydistillation} show that OPD achieves reasoning performance comparable to standard RL at a lower computational cost by providing dense supervision on the student's own rollouts. Inspired by such advancements, on-policy self-distillation (OPSD) is proposed to improve the model's performance via a self-teacher that leverages privileged information (e.g., ground truth)~\citep{zhao2026self, hubotter2026reinforcement}. Similarly, on-policy distillation can improve downstream task performance by incorporating demonstrations or additional feedback for continual learning~\citep{shenfeld2026self, he2026self}. To systematically understand these underlying dynamics, \citet{li2026rethinking} reveal that successful OPD requires a teacher that offers novel capabilities while maintaining compatible reasoning patterns with the student. Concurrent to our work, \citet{yang2026self} propose RLSD, which combines self-distillation and RLVR by using a privileged teacher to scale update magnitudes while environmental rewards govern update directions. Building on this idea, \citet{heakl2026cepo} and \citet{pan2026rlcsd} contrast teachers conditioned on correct and incorrect answers to derive token-level evidence, but leave uniformly failed groups without learning signals. Combining these complementary insights, we propose \ours{}: by unifying RLVR with OPSD, our approach jointly leverages sparse outcome signals and dense token-level supervision, yielding stronger reasoning performance without additional generation.
\section{Methodology}

\subsection{Preliminaries}
We consider the standard RLVR setting where an LLM-based policy $\pi_\theta$ is trained to maximize a verifiable reward $r(x, y)$ given a prompt $x$ and a generated response $y$. At each training iteration, the policy samples a group of $G$ rollouts $\{y_i\}_{i=1}^G \sim \pi_\theta(\cdot \mid x)$ for each prompt $x$. The reward function then scores each rollout, producing binary rewards $r_i \in \{0, 1\}$.

\textbf{GRPO.}
Group relative policy optimization~\citep{shao2024deepseekmath} replaces the learned value function with group statistics, $\hat{A}_i = \frac{r_i - \mu}{\sigma + \epsilon}$, where $\mu$ and $\sigma$ are the mean and standard deviation of the group rewards. Following Dr.~GRPO~\citep{liu2025understanding}, we drop $\sigma$, i.e., $\hat{A}_i = r_i - \mu$. Every token in $y_i$ shares this advantage, and the policy is trained with the policy-gradient surrogate~\citep{sutton1999policy}:
\begin{equation}
    \mathcal{L}_{\mathrm{GRPO}}(\theta) = -\mathbb{E}_{x \sim \mathcal{D},\, \{y_i\}_{i=1}^G \sim \pi_{\theta}(\cdot \mid x)} \left[\frac{1}{\sum_{i}|y_i|}\sum_{i=1}^{G}\sum_{t=1}^{|y_i|} \hat{A}_i \log \pi_\theta(y_{i,t} \mid x, y_{i,<t})\right],
    \label{eq:grpo_loss}
\end{equation}
whose gradient is the standard policy gradient; we omit the KL penalty to a reference policy.

\textbf{OPD.}
On-policy distillation~\citep{agarwal2024policy} instead lets a teacher $\pi_T$ re-score the student's own rollouts; minimizing the reverse KL on the sampled tokens~\citep{lu2025onpolicydistillation} gives the same form with a token-level advantage:
\begin{equation}
    \mathcal{L}_{\mathrm{OPD}}(\theta) = -\mathbb{E}_{x \sim \mathcal{D},\, \{y_i\}_{i=1}^G \sim \pi_{\theta}(\cdot \mid x)} \left[\frac{1}{\sum_{i}|y_i|}\sum_{i=1}^{G}\sum_{t=1}^{|y_i|} A^{\mathrm{OPD}}_{i,t} \log \pi_\theta(y_{i,t} \mid x, y_{i,<t})\right],
    \label{eq:opd_loss}
\end{equation}
where $A^{\mathrm{OPD}}_{i,t} = \mathrm{sg}\big[\log \pi_T(y_{i,t} \mid x, y_{i,<t}) - \log \pi_\theta(y_{i,t} \mid x, y_{i,<t})\big]$ and $\mathrm{sg}[\cdot]$ denotes stop-gradient, so each token's term is a sampled estimate of the reverse KL gradient $\nabla \mathbb{D}_{\mathrm{KL}}(\pi_\theta \,\|\, \pi_T)$.
OPD provides dense signals, but it depends on a compatible teacher and does not optimize the task reward directly.

\subsection{Self-Instructing Policy Optimization}

\ours{} keeps the policy-gradient objective of GRPO to directly optimize the task reward, but replaces its uniform advantage with a token-level advantage from a contrastive self-teacher, which re-evaluates each response under a correct and an incorrect privileged context.
We describe the details below.

\textbf{\ours{} Self-Teacher.}
\ours{} adopts the group-based sampling strategy of GRPO. Given a prompt $x$, the actor generates a group of $G$ candidate responses $\{y_i\}_{i=1}^G$. The reward function evaluates these rollouts to produce outcome rewards $r_i$. Together with the reference answer, these outcomes are then used to construct a pair of self-teacher contexts.

\begin{figure}[t]
    \centering
    \includegraphics[trim=2.9cm 3.4cm 2.7cm 3.4cm, clip, width=\textwidth]{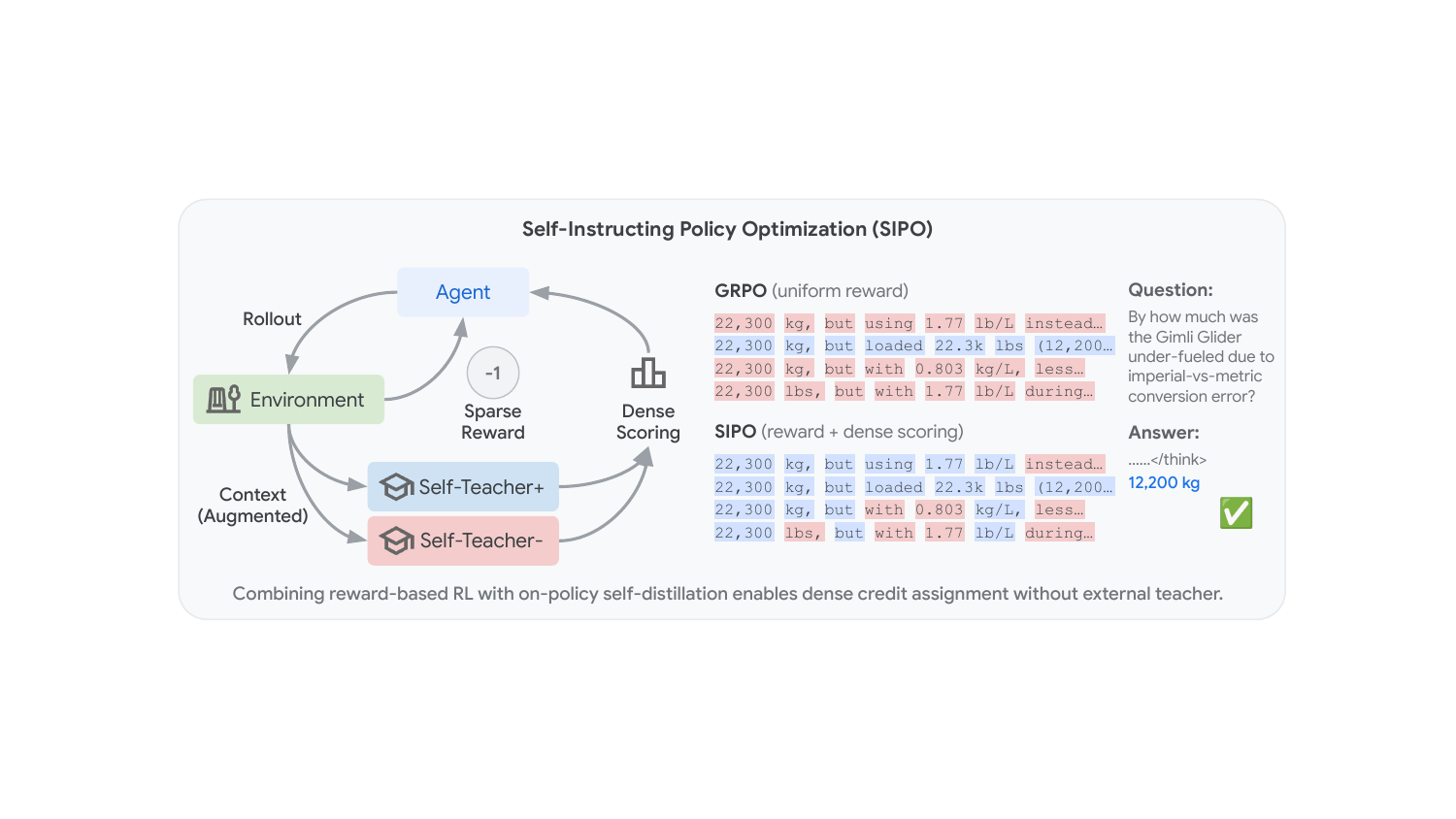}
    \caption{\textbf{Overview of \ours{}.} \ours{} combines sparse environment rewards with dense, contrastive self-teacher credit for fine-grained credit assignment. Left: the agent learns from outcome-level rewards and token-level scores from a self-teacher that contrasts a correct and an incorrect context. Right: unlike GRPO, which assigns a uniform advantage to all tokens, \ours{} aims to credit the reasoning steps (in blue) that lead to the correct answer.}
    \label{fig:method}
\end{figure}

\markblue{
For each prompt group, we split the rollouts into correct and incorrect ones, $\mathcal{S}^+(x) = \{j : r_j = 1\}$ and $\mathcal{S}^-(x) = \{j : r_j = 0\}$. For every rollout $y_i$, \ours{} builds two augmented teacher prompts, a positive prompt $x_i^+$ and a negative prompt $x_i^-$, sharing the same template and differing only in the privileged information placed before the original prompt $x$. When a reference answer $a^\star$ is available, the positive prompt contains it and the negative prompt contains an incorrect answer $a_i^-$:
\begin{equation}
    x_i^+ = \texttt{augment}(x, a^\star), \quad x_i^- = \texttt{augment}(x, a_i^-).
\end{equation}
For an incorrect rollout, $a_i^-$ is its own final answer as judged by the verifier; for a correct rollout, it is the most common incorrect answer in the group. Since $a_i^-$ depends on the outcome of the rollout itself, this construction is a form of hindsight relabeling. Incorrect rollouts without a valid judged answer, e.g., truncated responses, receive no teacher term, since substituting another answer from the group would contrast the reference with a mistake the rollout did not make. When no reference answer is available, as in code generation where the ground truth is a test suite, the two prompts instead contain a successful and a failed sibling rollout from the same group (a pair we also use on science QA whenever both siblings exist); rollouts without such a pair, including those in uniformly failed groups, receive no contrastive feedback (\Cref{sec:app_pairs}).
}

\markblue{
Conditioned on each augmented prompt, the self-teacher re-evaluates the \emph{same} response $y_i$:
\begin{equation}
    \pi_T(y_{i,t} \mid x_i^{\pm}, y_{i,<t})
    \triangleq
    \pi_{\bar{\theta}}(y_{i,t} \mid x_i^{\pm}, y_{i,<t}),
    \label{eq:teacher}
\end{equation}
where $\bar{\theta}$ denotes the teacher parameters (synchronized with the current policy every $K$ training steps).
The student is evaluated on the same response tokens under the original prompt-response pair $(x, y_i)$. Thus, \ours{} does not require additional generation: it only adds two teacher forward passes, one for each augmented prompt concatenated with the existing rollout. Since the teacher and student differ only in the prompt prefix, re-evaluating the shared response tokens avoids the need to resample, resulting in only a modest computational overhead compared to generation.
}

\textbf{\ours{} Objective.}
\markblue{
From the teacher evaluations, \ours{} computes a token-level contrastive evidence
\begin{equation}
    e_{i,t} = \mathrm{sg}\Big[\log \pi_T(y_{i,t} \mid x_i^+, y_{i,<t}) - \log \pi_T(y_{i,t} \mid x_i^-, y_{i,<t})\Big].
    \label{eq:evidence}
\end{equation}
A positive $e_{i,t}$ indicates that token $y_{i,t}$ is more likely when the teacher sees correct rather than incorrect information, and a negative $e_{i,t}$ indicates the opposite. We then add the clipped evidence to the group-relative advantage, and the \ours{} loss takes the same form as \Cref{eq:grpo_loss,eq:opd_loss}:
\begin{align}
    \mathcal{L}_{\mathrm{\ours}}(\theta) = - & \mathbb{E}_{x \sim \mathcal{D},\, \{y_i\}_{i=1}^G \sim \pi_{\theta}(\cdot \mid x)} \left[\frac{1}{\sum_{i}|y_i|}\sum_{i=1}^{G}\sum_{t=1}^{|y_i|} A_{i,t} \log \pi_\theta(y_{i,t} \mid x, y_{i,<t})\right], \nonumber \\
    & A_{i,t} = \hat{A}_i + \kappa \cdot \mathrm{clip}\big(e_{i,t},\, -\delta,\, \delta\big),
    \label{eq:sipo}
\end{align}
where $\kappa$ scales the teacher term, $\delta$ bounds the evidence per token, and $e_{i,t} = 0$ for rollouts without a contrastive pair. We use $\delta = 0.2$ and $\kappa \in \{0.1, 0.2, 0.5\}$ (\Cref{sec:app_exp}). \ours{} thus combines the sequence-level advantage of GRPO with a token-level advantage in the spirit of OPD, where the student log-probability in $A^{\mathrm{OPD}}_{i,t}$ is replaced by a second teacher evaluation. For clarity, we write all three losses in their on-policy form; in practice, we optimize the clipped importance-weighted surrogate~\citep{schulman2017proximal}. Because rewards are binary, the smallest nonzero $|\hat{A}_i|$ is $1/G$; with $\kappa\delta < 1/G$, the teacher term cannot flip the sign of $\hat{A}_i$ in groups with mixed outcomes (we additionally apply a sign-preserving projection as a safeguard; see \Cref{sec:app_proj}), so the reward alone determines the main direction of each update while the evidence redistributes credit across tokens. In a uniformly failed group, $\hat{A}_i = 0$ and $A_{i,t} = \kappa \cdot \mathrm{clip}(e_{i,t}, -\delta, \delta)$, so these rollouts, which GRPO discards, still receive a token-level signal indicating which tokens agree with the reference answer rather than with the rollout's own mistake.
}

In summary, the contrast between the two self-teacher evaluations provides a dense, token-level learning signal that addresses the fundamental limitations of sparse outcome rewards, even on uniformly failed groups. We expect tokens such as routine derivations or formatting to receive $e_{i,t} \approx 0$, while tokens specific to the correct or the incorrect answer receive positive or negative credit. By adding this evidence to the advantage, \ours{} is designed to indicate which intermediate steps moved the trajectory toward or away from the correct answer, while the reward keeps the direction of each update. Furthermore, because the incorrect answers are directly collected from the policy's own rollouts, the negative context tracks the mistakes the policy currently makes, keeping the contrast relevant as the policy improves. Therefore, the supervision signal evolves with every training iteration and requires neither an external teacher nor additional sampling, only two extra forward passes.

\section{Experiments}
\label{sec:exp}

\begin{table}[t]
\centering
\resizebox{\linewidth}{!}{%
\begin{tabular}{@{}lccccccc@{}}
\toprule
\multirow{2}{*}{}          & \multicolumn{3}{c}{Competition Math}          & \multicolumn{3}{c}{Standard Math}             & \multirow{2}{*}{Avg.} \\ \cmidrule(lr){2-4} \cmidrule(lr){5-7}
                           & AIME24        & AIME25        & AMC23         & MATH500       & Minerva       & Olympiad      &                 \\ \midrule
Qwen3-8B                   & 26.3          & 23.4          & 59.7          & 73.8          & 20.2          & 41.7          & 40.8            \\
+ GRPO                     &  \ul{54.9}    &  \ul{40.3}    & 84.7          & 84.6          & 29.4          & 50.1          &  \ul{57.3}      \\
+ SDPO                     & 33.4          & 25.5          & 63.8          & 77.6          & 25.4          & 43.5          & 44.9            \\
+ RLSD                     & 49.5          & 38.2          &  \ul{85.6}    &  \ul{84.8}    &  \ul{30.9}    &  \ul{51.2}    & 56.7            \\
+ \textbf{\ours{}}         & \textbf{56.8} & \textbf{41.4} & \textbf{87.8} & \textbf{85.8} & \textbf{32.0} & \textbf{53.0} & \textbf{59.4}   \\ \bottomrule
\end{tabular}
}
\vspace{3pt}
\caption{\textbf{Comparison of \ours{} and baselines on math reasoning benchmarks.} All methods train Qwen3-8B on DAPO-Math-17k; we report the best checkpoint with \texttt{avg@32} on AIME24 and AIME25, \texttt{avg@8} on AMC23, and \texttt{avg@1} on MATH500, Minerva Math (Minerva), and OlympiadBench (Olympiad). Best results are in \textbf{bold} and second-best results are \ul{underlined}.}
\label{tab:main_math}
\end{table}

We evaluate \ours{} in two scenarios: (1)~standard RLVR with binary rewards and a reference answer, and (2)~code generation, where correctness is determined by unit tests and no reference answer is available. The experimental settings are as follows:
\begin{itemize}[leftmargin=20pt]
    \item \emph{Datasets and Models.}
    We cover four domains with outcome-level correctness rewards: \textbf{math}, training on DAPO-Math-17k~\citep{yu2025dapo} and evaluating on AIME24, AIME25, AMC23, MATH-500~\citep{hendrycks2021measuring, lightman2023let}, Minerva Math and OlympiadBench~\citep{he2024olympiadbench}; \textbf{science QA} on SciKnowEval (Chemistry, Physics, Biology and Materials)~\citep{feng2024sciknoweval}; \textbf{tool use} on ToolAlpaca~\citep{tang2023toolalpaca}; and \textbf{code generation} on LiveCodeBench v6~\citep{jain2024livecodebench}, where unit tests determine the reward, with held-out IFEval~\citep{zhou2023instruction}, ArenaHard-v2~\citep{li2024crowdsourced}, and MMLU-Pro~\citep{wang2024mmlu} for generalization. We use Qwen3-8B~\citep{yang2025qwen3} on all tasks and Olmo3-7B-Instruct~\citep{olmo2025olmo} on science QA and tool use.
    \item \emph{Baselines.}
    We compare \ours{} against \textbf{GRPO}~\citep{shao2024deepseekmath} with asymmetric clipping~\citep{yu2025dapo}, \textbf{SDPO}~\citep{hubotter2026reinforcement}, which performs on-policy self-distillation without the policy-gradient term, \textbf{RLSD}~\citep{yang2026self}, which rescales the reward advantage with a privileged self-teacher, and, on code generation, SFT on self-teacher responses. All methods use \texttt{verl}~\citep{sheng2025hybridflow}, with details in \Cref{sec:app_exp}.
\end{itemize}

\begin{table}[t]
\centering
\resizebox{\linewidth}{!}{%
\begin{tabular}{@{}lccccccccccc@{}}
\toprule
\multirow{2}{*}{}          & \multicolumn{2}{c}{Chemistry} & \multicolumn{2}{c}{Physics}   & \multicolumn{2}{c}{Biology}   & \multicolumn{2}{c}{Materials} & \multicolumn{2}{c}{Tool use}  & \multirow{2}{*}{Avg.} \\ \cmidrule(lr){2-11}
                           & 1h            & 5h            & 1h            & 5h            & 1h            & 5h            & 1h            & 5h            & 1h            & 5h            &                          \\ \midrule
Qwen3-8B                   & \multicolumn{2}{c}{41.2}      & \multicolumn{2}{c}{59.2}      & \multicolumn{2}{c}{30.8}      & \multicolumn{2}{c}{58.9}      & \multicolumn{2}{c}{57.5}      & 49.5                     \\
+ GRPO                     & 63.3          & 63.4          & 63.6          & 63.6          &  \ul{49.8}    & 49.8          &  \ul{73.9}    & 74.1          & 60.2          & 65.7          & 62.7          \\
+ SDPO                     &  73.2         & \textbf{80.9} & 66.6          &  \ul{75.6}    & \textbf{50.6} &  \ul{56.8}    & 72.1          &  \ul{78.4}    & \textbf{68.0} & \textbf{68.5} &  \ul{69.1}    \\
+ RLSD                     &  \ul{73.2}    & 74.6          &  \ul{67.5}    & 70.3          & 44.0          & 44.0          & 66.7          & 69.0          & 62.4          & 62.5          & 63.4          \\
+ \textbf{\ours{}}         & \textbf{76.9} &  \ul{79.6}    & \textbf{74.6} & \textbf{79.2} & 46.8          & \textbf{58.7} & \textbf{77.1} & \textbf{79.5} &  \ul{64.6}    &  \ul{67.4}    & \textbf{70.4} \\ \bottomrule
\end{tabular}
}
\vspace{3pt}
\caption{\textbf{Comparison of \ours{} and baselines on science QA and tool use.} We report the best \texttt{avg@16} within 1h and 5h of wall-clock training. See \Cref{tab:app_grpo} for off-policy GRPO and \Cref{tab:app_olmo} for results with Olmo3-7B. Best results are in \textbf{bold} and second-best results are \ul{underlined}.}
\label{tab:main_results}
\end{table}

\begin{figure}[t]
    \centering
    \includegraphics[trim=1cm 0 1cm 0, clip, width=0.95\textwidth]{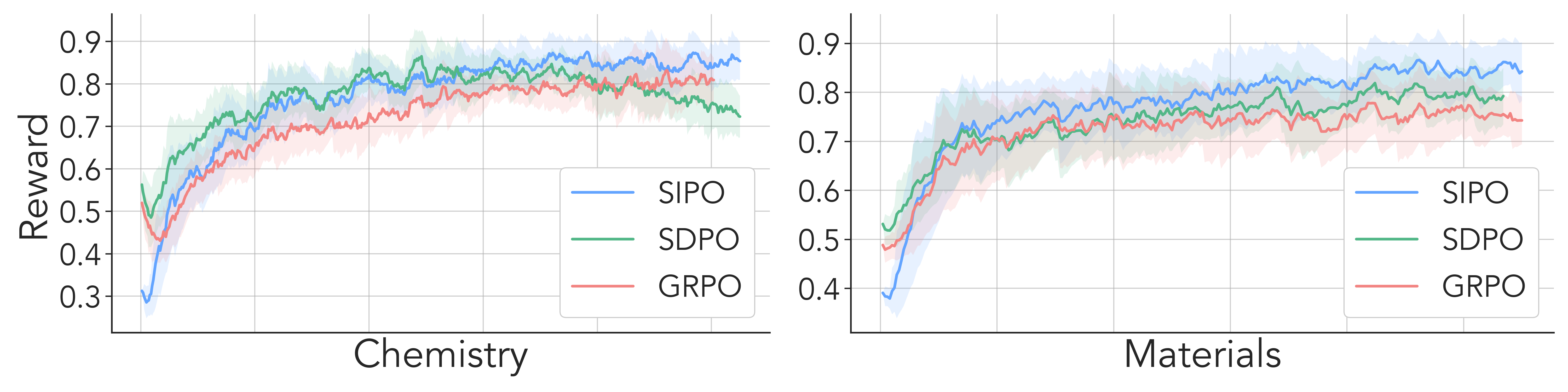}
    \caption{\textbf{Reward curves of \ours{}, SDPO, and GRPO on Chemistry and Materials.} \ours{} converges faster and reaches a higher final reward than both baselines by combining sparse outcome rewards with dense, token-level credit from the contrastive self-teacher.}
    \label{fig:reward_curves}
\end{figure}

\begin{table}[t]
\centering
\resizebox{\linewidth}{!}{%
\begin{tabular}{@{}lcccccc@{}}
\toprule
 & \multicolumn{1}{c}{Task} & \multicolumn{5}{c}{Holdout tasks} \\ \cmidrule(lr){2-2}\cmidrule(lr){3-7}
 & LCBv6 & IFEval & \makecell{\small ArenaHard-v2\\[-2pt]\small (hard prompt)}
 & \makecell{\small ArenaHard-v2\\[-2pt]\small (creative writing)} & MMLU-Pro & \makecell{\small Avg.\\[-2pt]\small (holdout)}\\ \midrule
Qwen3-8B         & 27.9          & \textbf{83.9} & \textbf{14.0} & \textbf{13.7} & 62.5          & \textbf{43.5} \\ \midrule
Self-teacher SFT & 42.7          &  \ul{83.7}    & 11.2          & 8.9           & 61.9          & 41.4          \\
GRPO             & 41.2          & 82.2          & 12.0          & 10.8          & 62.3          & 41.8          \\
SDPO             & 48.8          & 83.2          & 12.3          & 11.1          & \textbf{62.9} & 42.4          \\
RLSD             &  \ul{55.4}    & 83.0          & 12.3          & 11.8          & 62.5          & 42.4          \\
\textbf{\ours}   & \textbf{56.3} & 83.5          &  \ul{12.7}    &  \ul{12.0}    &  \ul{62.6}    &  \ul{42.7}    \\ \bottomrule
\end{tabular}
}
\vspace{3pt}
\caption{\textbf{Comparison of \ours{} and baselines on code generation and holdout tasks.} Holdout benchmarks only measure generalization beyond LCBv6, and self-teacher SFT trains on responses generated by the self-teacher. \ours{} achieves the highest LCBv6 accuracy while maintaining strong generalization. Best results are in \textbf{bold} and second-best results are \ul{underlined}.}
\label{tab:lcbv6_generalization}
\end{table}

\subsection{Results}

\begin{figure}[t]
    \centering
    \includegraphics[width=\textwidth]{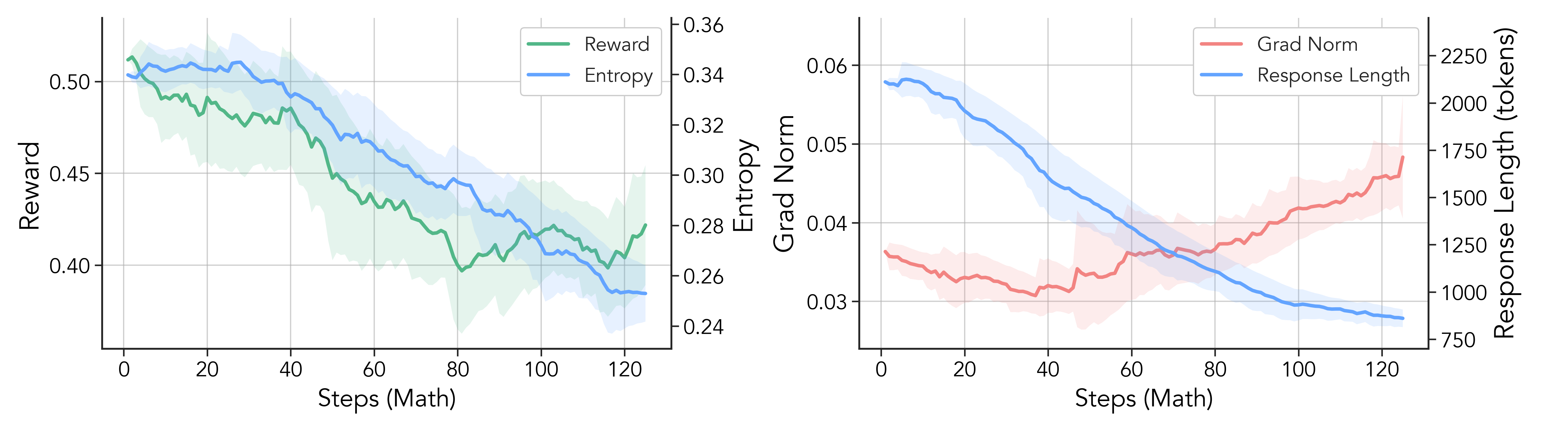}
    \caption{\textbf{Training dynamics of SDPO on DAPO-Math-17k with Qwen3-8B.} Left: training reward and entropy. Right: gradient norm and response length. Reward, entropy, and response length decrease steadily, while the gradient norm grows.}
    \label{fig:sdpo_math}
    \vspace{-10pt}
\end{figure}

\textbf{Math Results.}
\Cref{tab:main_math} summarizes results on the math reasoning benchmarks with Qwen3-8B. We make the following observations:
(1)~\emph{Overall performance}: \ours{} achieves the best result on all benchmarks and the highest average accuracy 59.4, improving over GRPO by 2.1\% and RLSD by 2.7\%.
(2)~\emph{Competition-level problems}: The gains hold on hard benchmarks, where \ours{} reaches 56.8 on AIME24 and 41.4 on AIME25, compared with 54.9 and 40.3 for GRPO. In contrast, RLSD falls below GRPO on both AIME benchmarks despite better results on standard benchmarks, suggesting that reweighting the reward advantage does not consistently help on long reasoning trajectories.
(3)~\emph{Comparison to self-distillation}: SDPO improves the base model by only 4.1\% on average and remains below other methods, consistent with the observation that a self-teacher alone struggles on complex reasoning (see \Cref{fig:sdpo_math}). By keeping the reward as the main learning signal and adding contrastive token-level credit, \ours{} benefits from self-distillation without inheriting this limitation, and it is the only self-distillation method that improves over GRPO on all six benchmarks.

\textbf{Reasoning Results.}
\Cref{tab:main_results} summarizes results on the reasoning benchmarks in science QA and tool use with Qwen3-8B (see \Cref{tab:app_olmo} for Olmo3-7B). Based on these results, we make the following observations:
(1)~\emph{Overall performance}: \ours{} achieves the highest average accuracy 70.4, representing an improvement over SDPO by 1.3\% and large gains over RLSD (by 7.0\%) and GRPO (by 7.7\%). Additionally, \ours{} secures the best result in 6 of the 10 settings, highlighting its effectiveness across scientific reasoning and tool use. The same trends hold with Olmo3-7B, where \ours{} improves over SDPO by 1.4\% and GRPO by 6.7\% on average (\Cref{tab:app_olmo}), and GRPO with four off-policy mini-batch steps per batch still falls well below \ours{} with both backbones (\Cref{tab:app_grpo}).
(2)~\emph{Training efficiency}: \ours{} learns faster than the baselines, with 1 hour of \ours{} often matching 5 hours of baseline training. For example, \ours{} reaches 76.9 on Chemistry at 1h, surpassing GRPO (63.4) and RLSD (74.6) at 5h. Similarly, \ours{} reaches 74.6 on Physics at 1h, exceeding 5h GRPO (63.6) and RLSD (70.3) and rivaling 5h SDPO before peaking at 79.2 at 5h. These results indicate that the self-teacher signal improves sample efficiency by converting reference answers and in-group mistakes into dense token-level supervision.
(3)~\emph{Comparison to reward-only and distillation-only baselines}: SDPO is competitive on several tasks and obtains the best result in 4 of the 10 settings, while RLSD improves over GRPO by only 0.7\% on average and falls behind it on Biology and Materials. However, \ours{} is the most consistent method on average, suggesting that combining reward-based policy gradients with on-policy self-distillation provides a stronger learning signal than either reward optimization or self-distillation alone.
(4)~\emph{SDPO on short versus long reasoning}: SDPO is close to \ours{} here but substantially worse on math (\Cref{tab:main_math}), likely because distilling toward an answer-aware teacher costs little on short responses but shortens the long reasoning needed for math.

\textbf{Code Generation Results.}
\Cref{tab:lcbv6_generalization} further evaluates whether the improvements from \ours{} generalize to code generation without a reference answer (LCBv6), as well as to held-out tasks (e.g., IFEval). We observe the following:
(1)~\emph{Learning without a reference answer}: 
On LiveCodeBench v6, \ours{} improves base model accuracy from 27.9 to 56.3, outperforming SFT on self-teacher responses, GRPO, SDPO and the strongest baseline RLSD (55.4). This demonstrates that contrasting successful and failed sibling rollouts effectively converts sparse code-verification outcomes into a much stronger learning signal. Unlike on math and science QA, uniformly failed groups receive no contrastive feedback on code generation, since no successful sibling is available; the gains therefore come from mixed groups alone, suggesting that a reference solution or execution feedback could further strengthen \ours{} on code.
(2)~\emph{Forgetting and generalization}: The holdout benchmarks are never used for training and only measure whether training on LCBv6 preserves the general capabilities of the model. \ours{} preserves holdout performance comparable to or better than the baseline methods. Its average holdout score is 42.7, close to the base model's 43.5 and higher than SFT, GRPO, SDPO and RLSD. In contrast, SFT improves LCBv6 but lowers the holdout average to 41.4, suggesting stronger forgetting behavior. Among all trained methods, \ours{} stays closest to the base model, losing only 0.8\% on average, so its gains on LCBv6 come with little forgetting.

\begin{wrapfigure}[18]{r}{0.46\textwidth}
    \centering
    \includegraphics[trim=0.5cm 0 1cm 0, clip, width=0.46\textwidth]{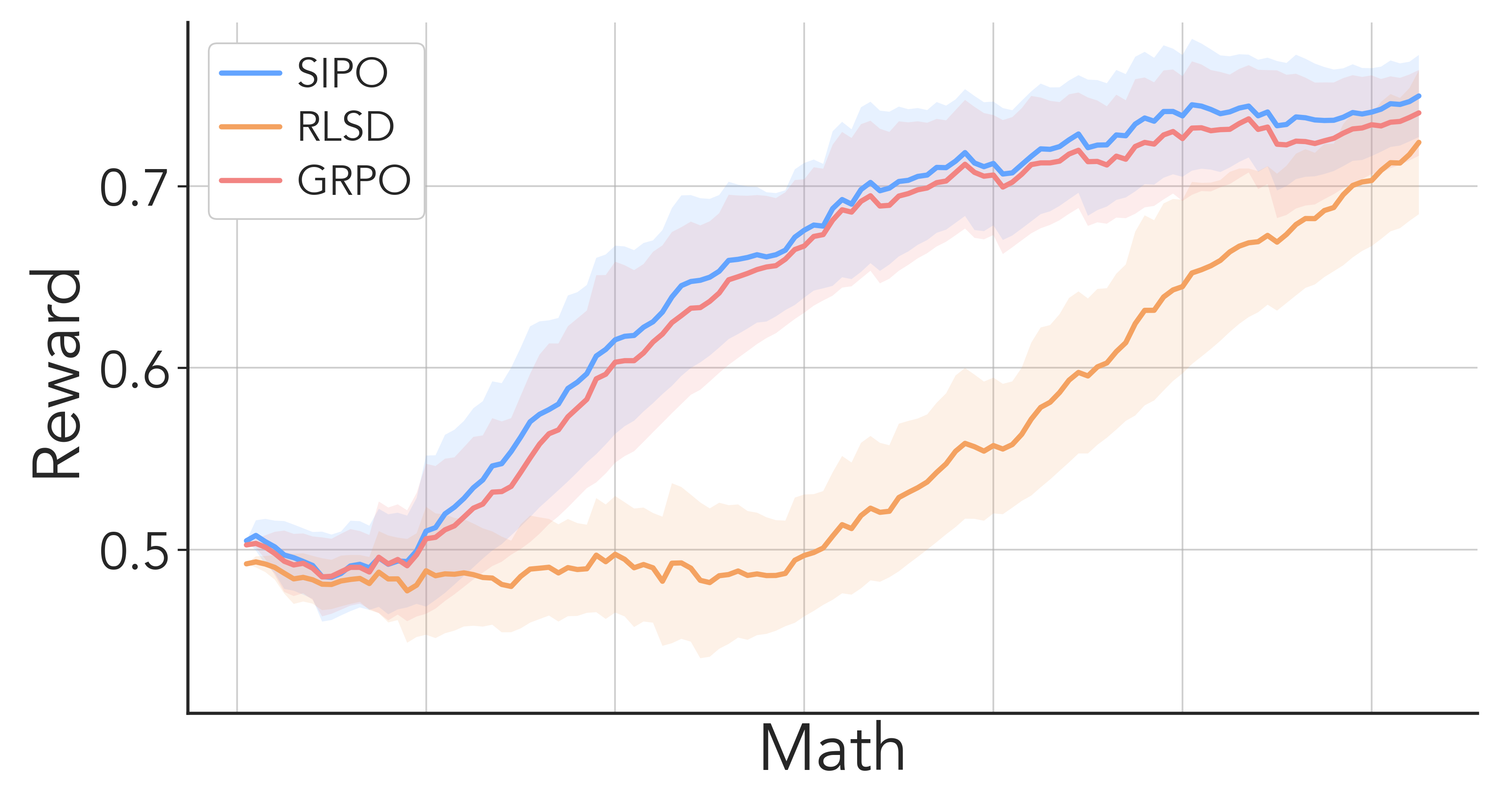}
    \caption{\textbf{Training reward of \ours{}, GRPO, and RLSD on Math with Qwen3-8B.} \ours{} improves from the start of training and stays above GRPO throughout, while RLSD stalls near its initial reward until its teacher weight decays to zero at step 50.}
    \label{fig:math_reward}
\end{wrapfigure}
\textbf{Learning Dynamics.}
Beyond final accuracy, we also examine the training dynamics of \ours{} to understand why combining reward-based optimization with a contrastive self-teacher yields a stronger learning signal. Based on \Cref{fig:reward_curves}, \Cref{fig:sdpo_math} and \Cref{fig:math_reward}, we make the following observations:
(1)~\emph{Reward trajectories}: \ours{} attains higher reward than both SDPO and GRPO throughout training and continues improving when the baselines begin to plateau. In Chemistry, SDPO peaks early and then steadily declines, GRPO plateaus, while \ours{} maintains a stable upward trajectory. We attribute this to the policy gradient term in \ours{}, which keeps optimizing toward verifiable outcomes even after the self-teacher signal saturates. 
(2)~\emph{Long reasoning on math}: On math (\Cref{fig:math_reward}), \ours{} and GRPO start from the same reward, after which \ours{} stays slightly above GRPO, whereas RLSD stalls while its teacher term is active and only begins to improve once its teacher weight $\lambda$ decays to zero at step 50, after which it reduces to GRPO. This suggests that contrastive credit adds to the reward signal without slowing learning, whereas reweighting the advantage with a self-teacher may delay progress on long trajectories.
(3)~\emph{Self-distillation collapses on long reasoning}: On math, the reward, entropy, and response length of SDPO steadily decrease, with responses shrinking below 1{,}000 tokens, while the gradient norm grows (\Cref{fig:sdpo_math}). We hypothesize that an answer-aware self-teacher favors continuations that head directly to the answer, so distilling toward it cuts the long reasoning that math requires, whereas on the short responses of science QA and tool use this shortcut costs little.
(4)~\emph{Why \ours{} avoids this failure}: Both teacher contexts carry an answer, so the preference for short continuations that head directly to the answer is expected to appear on both sides and largely cancel in the contrast. The remaining evidence only reflects which tokens are specific to the correct or the incorrect answer, while the reward still sets the update direction, so \ours{} keeps the long reasoning that math requires. Consistent with this, within the same math run, the single-teacher evidence $\log\pi_T(y_{i,t}\mid x_i^+,y_{i,<t})-\log\pi_\theta(y_{i,t}\mid x,y_{i,<t})$ has a near-zero median but a consistently negative mean ($-0.002$ to $-0.004$) that grows during training, indicating a tail of heavily penalized tokens, whereas the contrastive evidence $e_{i,t}$ fluctuates around zero ($|\text{mean}| < 0.001$) without a consistent sign.

\begin{table}[t]
\centering
\resizebox{\linewidth}{!}{%
\begin{tabular}{@{}lccccccc@{}}
\toprule
\multirow{2}{*}{}          & \multicolumn{3}{c}{\textbf{Competition Math}}      & \multicolumn{3}{c}{\textbf{Standard Math}}         & \multirow{2}{*}{Avg.} \\ \cmidrule(lr){2-4} \cmidrule(lr){5-7}
                           & AIME24          & AIME25          & AMC23           & MATH500         & Minerva         & Olympiad        &                 \\ \midrule
Qwen3-8B                   & 26.3            & 23.4            & 59.7            & 73.8            & 20.2            & 41.7            & 40.8            \\
+ GRPO                     & 54.9            &  \ul{40.3}      & 84.7            & 84.6            & 29.4            & 50.1            & 57.3            \\
+ GRPO w/ analytic KL      & 25.6            & 18.2            & 66.5            & 76.2            & 27.5            & 41.6            & 42.6            \\
+ RLSD                     & 49.5            & 38.2            & 85.6            &  \ul{84.8}      & 30.9            & 51.2            & 56.7            \\
+ RLSD w/ contrastive      &  \ul{56.7}      & 39.8            &  \ul{86.8}      & 83.2            &  \ul{31.2}      &  \ul{51.9}      &  \ul{58.3}      \\
+ \textbf{\ours{}}         & \textbf{56.8}   & \textbf{41.4}   & \textbf{87.8}   & \textbf{85.8}   & \textbf{32.0}   & \textbf{53.0}   & \textbf{59.4}   \\ \bottomrule
\end{tabular}
}
\vspace{3pt}
\caption{\textbf{Ablation of how the self-teacher is used on math with Qwen3-8B.} \textbf{GRPO w/ analytic KL} adds an analytic reverse KL toward a privileged self-teacher. \textbf{RLSD} rescales the advantage with token-level evidence from a self-teacher conditioned on the reference answer. \textbf{RLSD w/ contrastive} replaces this evidence with the contrast between a correct and an incorrect teacher context. \ours{} adds the contrastive evidence to the advantage of every rollout, including uniformly failed groups. Best results are in \textbf{bold} and second-best results are \ul{underlined}.}
\label{tab:ablation_math}
\end{table}

\textbf{Ablation Studies.}
\Cref{tab:ablation_math} examines how the self-teacher should enter the update on math. We make the following observations:
(1)~\emph{Distribution-level distillation collapses on long reasoning}: Adding an analytic KL toward a single self-teacher to GRPO decreases the reward throughout training, and its best checkpoint, near the start of training, reaches only 42.6 on average, 14.7\% below GRPO and barely above the base model. As with SDPO (\Cref{fig:sdpo_math}), matching a single answer-conditioned teacher likely removes the long reasoning required on math. On science QA and tool use, whose responses are short, the same variant remains competitive (71.2 vs.\ 70.4 for \ours{} with Qwen3-8B; \Cref{tab:app_analytic}), so this failure is specific to long reasoning.
(2)~\emph{A single teacher gives weak token-level credit}: RLSD, which uses a single teacher only to rescale the reward advantage, avoids this collapse but stays slightly below GRPO (56.7 vs.\ 57.3), even though its teacher weight decays to zero after 50 steps, and learns more slowly while the teacher is active (\Cref{fig:math_reward}).
(3)~\emph{The contrastive teacher is the key ingredient}: Replacing the evidence of RLSD with the contrast between a correct and an incorrect teacher context, which also allows keeping the teacher weight at 0.5 throughout training instead of decaying it to zero, raises the average from 56.7 to 58.3, surpassing GRPO, with the largest gain on AIME24. This suggests that the benefit comes from the contrast rather than from token-level reweighting alone.
(4)~\emph{Learning from every rollout}: Adding the contrastive evidence to the advantage of every rollout, including uniformly failed groups (with the negative contexts in \Cref{sec:app_pairs}), further raises the average to 59.4, with gains on all six benchmarks.
\section{Conclusion}
In this paper, we introduced \ours{}, which unifies reinforcement learning with on-policy self-distillation through a single token-level advantage. Instead of distilling toward a single self-teacher, \ours{} contrasts the teacher's evaluations under a correct and an incorrect privileged context, so that shared biases largely cancel and even uniformly failed groups receive dense credit. Across math, science QA, tool use and code generation tasks, \ours{} outperforms RL and self-distillation baselines, remains stable on long math reasoning where self-distillation collapses, and requires no external teacher or additional generation. Moreover, \ours{} forgets little on holdout evaluation and adds only teacher forward passes, making it a drop-in replacement for GRPO when environmental feedback is available.


\bibliographystyle{iclr2027_conference}
\bibliography{reference}      


\clearpage
\appendix
\section{Method Details}
\label{sec:app_method}

\subsection{Construction of Contrastive Pairs}
\label{sec:app_pairs}
In the main text, the two teacher prompts pair the reference answer with an incorrect answer $a_i^-$, and \Cref{fig:prompt} shows their template. This section details the remaining cases. When no reference answer is available (e.g., in code generation), the two teacher prompts instead contain a successful sibling $y_j^+$ and a failed sibling $y_k^-$ from the same group:
\begin{equation}
    x_i^+ = \texttt{augment}(x, y_j^+), \quad x_i^- = \texttt{augment}(x, y_k^-), \quad j \in \mathcal{S}^+(x),\; k \in \mathcal{S}^-(x).
\end{equation}
Both siblings differ from $y_i$, and we also use this pair on science QA whenever both exist. Unlike \citet{hubotter2026reinforcement}, we do not add environment feedback to the teacher prompts, so that the two contexts differ only in the correct versus incorrect information.

A rollout receives contrastive feedback only if both of its teacher prompts can be constructed, which excludes three cases. First, uniformly correct groups contain no incorrect answer or failed sibling to serve as the negative context, and we do not construct an artificial one, so these rollouts receive neither a nonzero advantage nor a teacher signal. Second, some incorrect rollouts have no valid judged answer, e.g., a math response that is truncated or places its boxed answer before further text, or a multiple-choice response without an answer tag. Since the negative context of an incorrect rollout is its own answer, we skip these rollouts instead of substituting another answer from the group; in preliminary experiments, substituting the most common incorrect answer of the group led to a growing positive advantage on uniformly failed groups as the fraction of truncated responses increased. Third, when no reference answer is available (code generation), the contrast requires both a successful and a failed sibling, so rollouts in uniformly failed groups receive no contrastive feedback. On science QA, which provides reference answers, such groups fall back to the answer-based pair.

\begin{figure}[t]
\begin{prompt}{Prompt for \ours}
\texttt{<|im\_start|>user}

\texttt{Here is a reference solution:}

\texttt{<reference answer $a^\star$ | incorrect answer $a_i^-$>}

\texttt{ }

\texttt{After understanding the reference solution, please try to solve this problem using your own approach below.}

\texttt{---}

\texttt{<question><|im\_end|>}

\texttt{<|im\_start|>assistant}

\end{prompt}
\caption{\textbf{Prompt template for the \ours{} self-teacher.} The positive prompt $x_i^+$ contains the reference answer and the negative prompt $x_i^-$ the incorrect answer $a_i^-$. Without a reference answer (code generation), the slot holds a successful or a failed sibling rollout (\Cref{sec:app_pairs}).}
\label{fig:prompt}
\end{figure}

\subsection{Sign-Preserving Projection}
\label{sec:app_proj}
To guarantee that the teacher term never reverses the direction set by the reward, we apply a sign-preserving projection to the token-level advantage in \Cref{eq:sipo}:
\begin{equation}
    A_{i,t} = \Pi_{\hat{A}_i}\Big(\hat{A}_i + \kappa \cdot \mathrm{clip}\big(e_{i,t},\, -\delta,\, \delta\big)\Big),
    \qquad
    \Pi_{\hat{A}}(z) =
    \begin{cases}
        \max(z, 0) & \hat{A} > 0, \\
        \min(z, 0) & \hat{A} < 0, \\
        z & \hat{A} = 0.
    \end{cases}
\end{equation}
With binary rewards, the smallest nonzero $|\hat{A}_i|$ is $1/G$, so the projection is inactive whenever $\kappa\delta < 1/G$, which holds in all our settings ($\kappa\delta \le 0.1 < 1/8$). Rollouts in uniformly failed groups ($\hat{A}_i = 0$) are left unchanged, so their advantage keeps the sign of the evidence.

\section{Experiment Settings}
\label{sec:app_exp}

We evaluate \ours{} in two settings: (1)~standard RLVR environments, where feedback is limited to binary rewards and a reference answer is available, and (2)~a code generation environment, where correctness is determined by unit tests and no reference answer is available. In standard RLVR tasks, \ours{} contrasts the reference answer with incorrect answers from the same group. By comparing its evaluations under the two contexts, the self-teacher is designed to identify which tokens support the correct answer and to provide dense credit assignment. For code generation, outputs are executed against training unit tests, with separate test cases held out for evaluation. The environment returns a binary reward, which is one only if all training tests pass. Since no reference answer exists, \ours{} contrasts a successful and a failed same-question rollout (\Cref{sec:app_pairs}), converting sparse code-verification outcomes into dense token-level learning signals.

\paragraph{Datasets and Models.}
We evaluate \ours{} and baseline methods across four benchmark families: \textbf{math}, \textbf{science QA}, \textbf{tool use}, and \textbf{code generation}. For \textbf{math}, we train on the deduplicated English subset of DAPO-Math-17k~\citep{yu2025dapo} (14{,}116 problems) and evaluate on AIME24, AIME25, AMC23, MATH-500~\citep{hendrycks2021measuring, lightman2023let}, Minerva Math, and OlympiadBench~\citep{he2024olympiadbench}, reporting \texttt{avg@32} on AIME24 and AIME25, \texttt{avg@8} on AMC23, and \texttt{avg@1} on the remaining benchmarks. For \textbf{science QA}, we use undergraduate-level reasoning subsets from SciKnowEval~\citep{feng2024sciknoweval}, including Chemistry, Physics, Biology, and Materials Science. For \textbf{tool use}, we use ToolAlpaca~\citep{tang2023toolalpaca}, where the model must map a tool-API specification and user request to the correct tool call. For \textbf{code generation}, we use questions from LiveCodeBench v6~\citep{jain2024livecodebench}, where generated solutions are evaluated by unit tests. For science QA, tool use, and code generation, we construct a train-test split to measure in-domain generalization. For LiveCodeBench v6, we additionally split the available unit tests into training tests used for the reward and held-out tests used for evaluation. We follow the data preprocessing pipeline of~\citet{hubotter2026reinforcement} and adopt their reward formulations and evaluation protocols to ensure direct comparability. We use Qwen3-8B~\citep{yang2025qwen3} on all tasks and additionally Olmo3-7B-Instruct~\citep{olmo2025olmo} on science QA and tool use.

\paragraph{Baselines and Implementation Details.}
We compare \ours{} against an improved variant of \textbf{GRPO}~\citep{shao2024deepseekmath}, which incorporates recent modifications such as asymmetric clipping~\citep{yu2025dapo}. All methods take a single gradient step per generation batch, i.e., they are trained on-policy, except on LCBv6, where all methods take four mini-batch steps per batch following the GRPO setting of \citet{hubotter2026reinforcement}; we additionally report GRPO with four off-policy mini-batch steps per batch on science QA and tool use in \Cref{tab:app_grpo}. To isolate the contribution of reward-based optimization, we further compare against \textbf{SDPO}~\citep{hubotter2026reinforcement}, which performs on-policy self-distillation without the policy gradient term. We also compare against \textbf{RLSD}~\citep{yang2026self}, which uses a privileged self-teacher to rescale the magnitude of the reward advantage, with a teacher weight that decays linearly from 0.5 to zero over the first 50 steps, as in the original setting. For all methods, we use the \texttt{verl} implementation~\citep{sheng2025hybridflow}, perform a hyperparameter sweep, and report results for the configurations that achieve the best performance across target tasks. For \ours{}, we use $\delta = 0.2$ and tune $\kappa \in \{0.1, 0.2, 0.5\}$. We use a learning rate of $10^{-5}$ for science QA and tool use, and $10^{-6}$ for math and LCBv6. Across all settings, we use rollout group size $G=8$ and token-level rollout importance correction; the train batch size is 32 prompts, except for math, where it is 128. The self-teacher is a periodic snapshot of the policy, synchronized every 10 training steps. Environment feedback is not used in the teacher prompts; on LCBv6, the teacher contexts are sibling rollouts (\Cref{sec:app_pairs}). For science QA and tool use, we report \texttt{avg@16} relative to wall-clock training time, with each run allocated up to 12 hours; for math, we train for 125 steps and report the best checkpoint. The prompt template is shown in \Cref{fig:prompt}.

\section{Additional Results}
\label{sec:app_res}

\subsection{Results with Olmo3-7B}
\begin{table}[t]
\centering
\resizebox{\linewidth}{!}{%
\begin{tabular}{@{}lccccccccccc@{}}
\toprule
\multirow{2}{*}{}          & \multicolumn{2}{c}{Chemistry} & \multicolumn{2}{c}{Physics}   & \multicolumn{2}{c}{Biology}   & \multicolumn{2}{c}{Materials} & \multicolumn{2}{c}{Tool use}  & \multirow{2}{*}{Avg.} \\ \cmidrule(lr){2-11}
                           & 1h            & 5h            & 1h            & 5h            & 1h            & 5h            & 1h            & 5h            & 1h            & 5h            &                          \\ \midrule
Olmo3-7B                   & \multicolumn{2}{c}{22.8}      & \multicolumn{2}{c}{37.7}      & \multicolumn{2}{c}{16.2}      & \multicolumn{2}{c}{36.7}      & \multicolumn{2}{c}{39.3}      & 30.5                     \\
+ GRPO                     & 51.4          & 57.5          &  \ul{62.7}    & 62.7          &  \ul{49.8}    & 49.8          & 73.3          & 73.5          & 56.8          & 60.6          & 59.8          \\
+ SDPO                     &  \ul{68.0}    & \textbf{80.0} & 59.9          &  \ul{66.1}    & 48.0          &  \ul{52.8}    &  \ul{73.7}    & \textbf{79.1} & \textbf{60.8} &  \ul{62.1}    &  \ul{65.1}    \\
+ \textbf{\ours{}}         & \textbf{71.2} &  \ul{75.9}    & \textbf{66.0} & \textbf{66.4} & \textbf{53.1} & \textbf{55.0} & \textbf{76.9} &  \ul{77.0}    &  \ul{58.8}    & \textbf{64.7} & \textbf{66.5} \\ \bottomrule
\end{tabular}
}
\vspace{3pt}
\caption{\textbf{Comparison of \ours{} and baselines on science QA and tool use with Olmo3-7B.} The setting follows \Cref{tab:main_results}. Best results are in \textbf{bold} and second-best results are \ul{underlined}.}
\label{tab:app_olmo}
\end{table}

\Cref{tab:app_olmo} reports the science QA and tool use results with Olmo3-7B.
The trends match those with Qwen3-8B: \ours{} achieves the highest average accuracy (66.5), improving over SDPO by 1.4\% and GRPO by 6.7\%, and obtains the best result in 7 of the 10 settings. SDPO remains competitive on Chemistry and Materials at 5h, consistent with its behavior on short responses discussed in \Cref{sec:exp}.

\subsection{Off-Policy GRPO}
\begin{table}[t]
\centering
\resizebox{\linewidth}{!}{%
\begin{tabular}{@{}lccccccccccc@{}}
\toprule
\multirow{2}{*}{}          & \multicolumn{2}{c}{Chemistry} & \multicolumn{2}{c}{Physics}   & \multicolumn{2}{c}{Biology}   & \multicolumn{2}{c}{Materials} & \multicolumn{2}{c}{Tool use}  & \multirow{2}{*}{Avg.} \\ \cmidrule(lr){2-11}
                           & 1h            & 5h            & 1h            & 5h            & 1h            & 5h            & 1h            & 5h            & 1h            & 5h            &                          \\ \midrule
Qwen3-8B                   & \multicolumn{2}{c}{41.2}      & \multicolumn{2}{c}{59.2}      & \multicolumn{2}{c}{30.8}      & \multicolumn{2}{c}{58.9}      & \multicolumn{2}{c}{57.5}      & 49.5                     \\
+ GRPO (off-policy)        & \textbf{65.9} & \textbf{74.5} & \textbf{63.8} & \textbf{72.7} & 35.1          & \textbf{59.9} & \textbf{74.3} & \textbf{77.1} & \textbf{64.9} & \textbf{67.7} & \textbf{65.6} \\
+ GRPO (on-policy)         & 63.3          & 63.4          & 63.6          & 63.6          & \textbf{49.8} & 49.8          & 73.9          & 74.1          & 60.2          & 65.7          & 62.7          \\ \midrule
Olmo3-7B                   & \multicolumn{2}{c}{22.8}      & \multicolumn{2}{c}{37.7}      & \multicolumn{2}{c}{16.2}      & \multicolumn{2}{c}{36.7}      & \multicolumn{2}{c}{39.3}      & 30.5                     \\
+ GRPO (off-policy)        & 39.7          & 56.7          & 55.3          & \textbf{63.3} & 35.6          & \textbf{55.8} & 70.9          & \textbf{75.0} & 56.4          & \textbf{65.0} & 57.4          \\
+ GRPO (on-policy)         & \textbf{51.4} & \textbf{57.5} & \textbf{62.7} & 62.7          & \textbf{49.8} & 49.8          & \textbf{73.3} & 73.5          & \textbf{56.8} & 60.6          & \textbf{59.8} \\ \bottomrule
\end{tabular}
}
\vspace{3pt}
\caption{\textbf{GRPO with off-policy and on-policy updates.} GRPO (off-policy) takes four mini-batch gradient steps per generation batch, while GRPO (on-policy) takes one step, matching the setting of all methods in \Cref{tab:main_results}. We report the best \texttt{avg@16} within 1h and 5h of wall-clock training. The better result for each base model is in \textbf{bold}.}
\label{tab:app_grpo}
\end{table}

\Cref{tab:app_grpo} compares GRPO with four off-policy mini-batch steps per generation batch against its on-policy variant, which is the GRPO baseline reported in the main text. Off-policy updates help GRPO with Qwen3-8B (65.6 vs.\ 62.7 on average) but not with Olmo3-7B (57.4 vs.\ 59.8), and both variants remain well below \ours{} (70.4 and 66.5), so the gains of \ours{} do not stem from restricting GRPO to on-policy updates.

\subsection{Reward Curves on Physics and Biology}
\begin{figure}[t]
    \centering
    \includegraphics[trim=1cm 0 1cm 0, clip, width=0.95\textwidth]{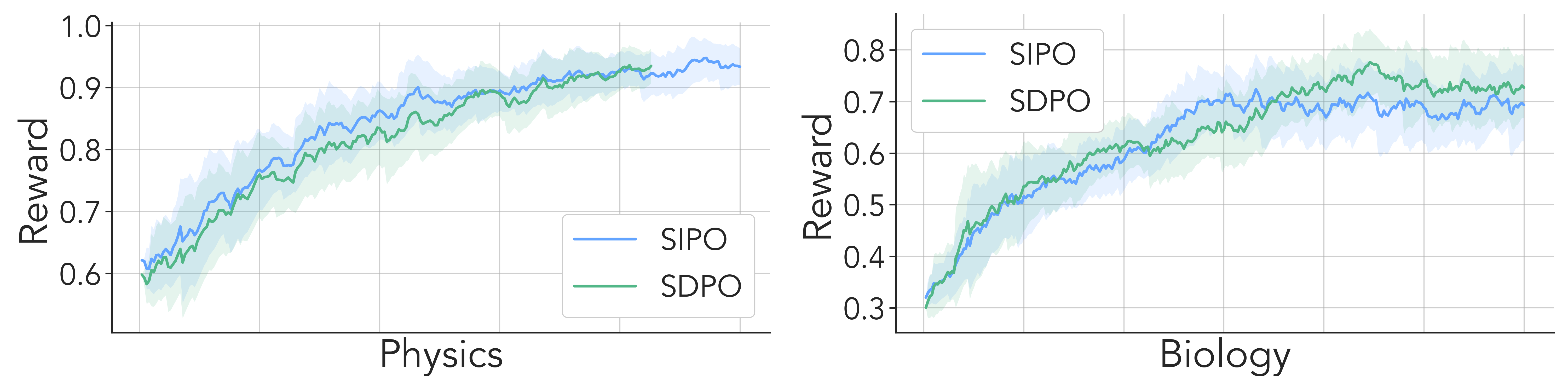}
    \caption{\textbf{Reward curves of \ours{} and SDPO on Physics and Biology with Qwen3-8B.} \ours{} reaches higher reward earlier in training and maintains a more stable upward trajectory.}
    \label{fig:reward_curves_2}
\end{figure}

\Cref{fig:reward_curves_2} shows the reward curves of \ours{} and SDPO on Physics and Biology with Qwen3-8B. \ours{} reaches higher reward earlier in training and keeps improving. On Physics, this translates into 74.6 at 1h versus 66.6 for SDPO (\Cref{tab:main_results}); on Biology, SDPO is slightly ahead at 1h (50.6 vs.\ 46.8), but \ours{} overtakes it at 5h (58.7 vs.\ 56.8).

\subsection{Comparison with the Analytic KL Variant}
\label{sec:app_analytic}
\begin{table}[t]
\centering
\resizebox{\linewidth}{!}{%
\begin{tabular}{@{}lccccccccccc@{}}
\toprule
\multirow{2}{*}{}                & \multicolumn{2}{c}{Chemistry} & \multicolumn{2}{c}{Physics}   & \multicolumn{2}{c}{Biology}   & \multicolumn{2}{c}{Materials} & \multicolumn{2}{c}{Tool use}  & \multirow{2}{*}{Avg.} \\ \cmidrule(lr){2-11}
                                 & 1h            & 5h            & 1h            & 5h            & 1h            & 5h            & 1h            & 5h            & 1h            & 5h            &                          \\ \midrule
\multicolumn{12}{@{}l}{\emph{Qwen3-8B}} \\
\ours{} (analytic KL)            & \textbf{78.8} & 79.2          & 73.8          & \textbf{81.5} & \textbf{55.3} & 56.0          & 74.0          & 76.4          & \textbf{68.3} & \textbf{68.3} & \textbf{71.2} \\
\ours{} (contrastive)            & 76.9          & \textbf{79.6} & \textbf{74.6} & 79.2          & 46.8          & \textbf{58.7} & \textbf{77.1} & \textbf{79.5} & 64.6          & 67.4          & 70.4          \\ \midrule
\multicolumn{12}{@{}l}{\emph{Olmo3-7B}} \\
\ours{} (analytic KL)            & \textbf{79.1} & \textbf{81.1} & \textbf{68.1} & \textbf{73.3} & \textbf{53.5} & 54.0          & \textbf{77.6} & \textbf{79.3} & \textbf{64.0} & \textbf{67.3} & \textbf{69.7} \\
\ours{} (contrastive)            & 71.2          & 75.9          & 66.0          & 66.4          & 53.1          & \textbf{55.0} & 76.9          & 77.0          & 58.8          & 64.7          & 66.5          \\ \bottomrule
\end{tabular}
}
\vspace{3pt}
\caption{\textbf{Analytic KL versus contrastive token-level advantage on science QA and tool use.} \ours{} (analytic KL) adds an analytic KL divergence toward a self-teacher conditioned on successful sibling rollouts to the GRPO objective; \ours{} (contrastive) is the final method. The setting follows \Cref{tab:main_results}. The better result for each base model is in \textbf{bold}.}
\label{tab:app_analytic}
\end{table}

\Cref{tab:app_analytic} compares \ours{} with an earlier variant that, instead of the contrastive token-level advantage, adds an analytic KL divergence between the student and a self-teacher conditioned on successful sibling rollouts to the GRPO objective (denoted GRPO w/ analytic KL in \Cref{tab:ablation_math}). On science QA and tool use, whose responses are relatively short, the analytic KL variant performs comparably or better (71.2 vs.\ 70.4 with Qwen3-8B and 69.7 vs.\ 66.5 with Olmo3-7B). On math, however, where responses span thousands of tokens, the analytic KL variant collapses during training and ends barely above the base model (\Cref{tab:ablation_math}), which we attribute to the overconfident self-teacher imposing excessive penalties over long reasoning trajectories. We therefore adopt the contrastive formulation, which remains stable on both short and long reasoning tasks.

\begin{table}[t]
\centering
\resizebox{\linewidth}{!}{%
\begin{tabular}{@{}lcccccc@{}}
\toprule
 & \multicolumn{1}{c}{Task} & \multicolumn{5}{c}{Holdout tasks} \\ \cmidrule(lr){2-2}\cmidrule(lr){3-7}
 & LCBv6 & IFEval & \makecell{\small ArenaHard-v2\\[-2pt]\small (hard prompt)}
 & \makecell{\small ArenaHard-v2\\[-2pt]\small (creative writing)} & MMLU-Pro & \makecell{\small Avg.\\[-2pt]\small (holdout)}\\ \midrule
\ours{} (analytic KL)   & \textbf{58.5} & \textbf{83.8} & \textbf{13.5} & \textbf{13.9} & \textbf{63.7} & \textbf{43.7} \\
\ours{} (contrastive)   & 56.3          & 83.5          & 12.7          & 12.0          & 62.6          & 42.7          \\ \bottomrule
\end{tabular}
}
\vspace{3pt}
\caption{\textbf{Analytic KL versus contrastive token-level advantage on LCBv6 and holdout tasks with Qwen3-8B.} The setting follows \Cref{tab:lcbv6_generalization}. The better result is in \textbf{bold}.}
\label{tab:app_analytic_code}
\end{table}

\Cref{tab:app_analytic_code} shows the same comparison on code generation, where the analytic KL variant, whose self-teacher is additionally conditioned on execution feedback, also performs slightly better (58.5 vs.\ 56.3 on LCBv6 and 43.7 vs.\ 42.7 on the holdout average). As on science QA, generated programs are compact compared with long mathematical derivations, and execution feedback tells the self-teacher where a program fails, so matching the teacher's full distribution likely remains a useful dense signal rather than a shortcut that removes necessary reasoning.

\begin{figure}[t]
    \centering
    \includegraphics[trim=0 0 0 0, clip, width=0.95\textwidth]{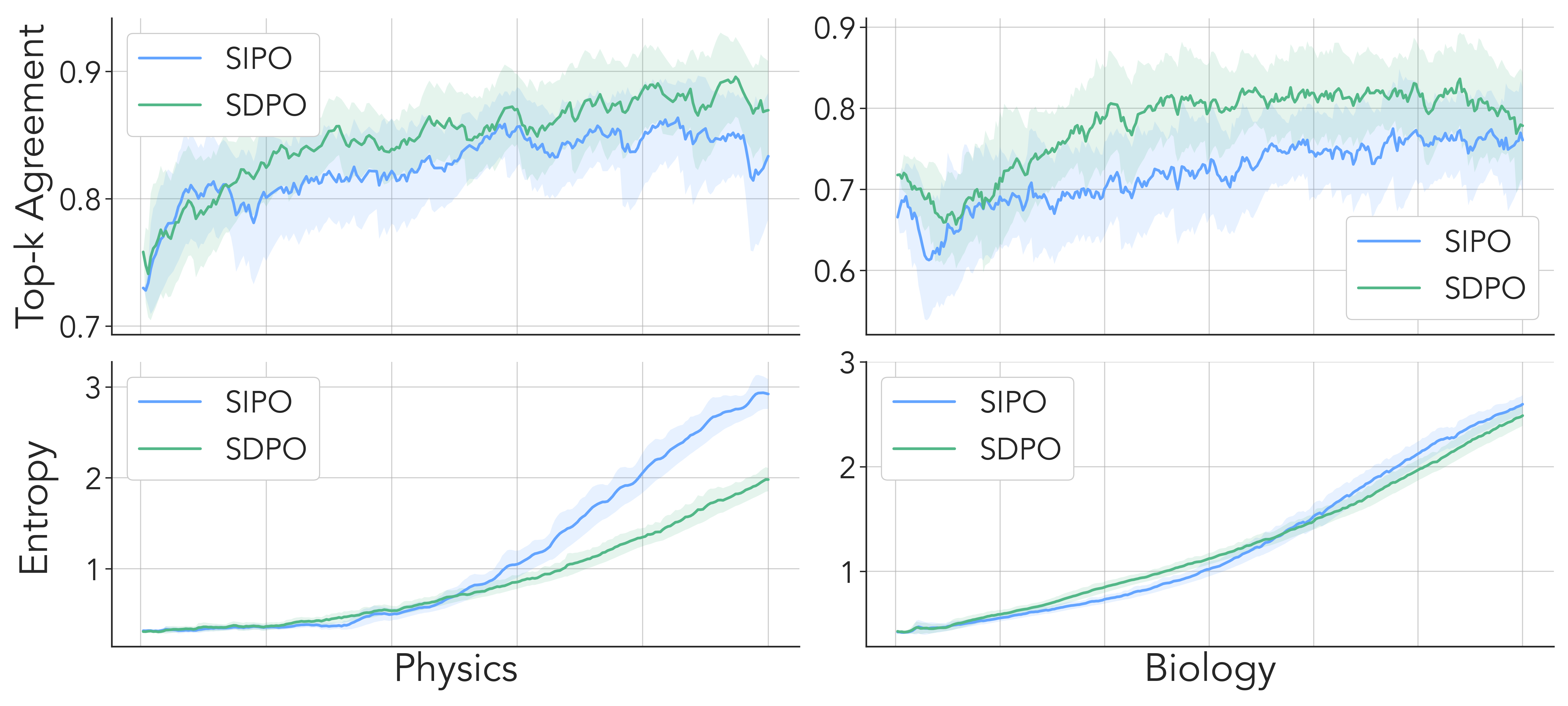}
    \caption{\textbf{Top-$k$ token agreement and entropy of the analytic KL variant of \ours{} versus SDPO on Physics and Biology.} The analytic KL variant exhibits slightly lower top-$k$ agreement, suggesting less over-optimization and greater diversity, while maintaining comparable generation entropy.}
    \label{fig:topk_entropy}
\end{figure}

\Cref{fig:topk_entropy} further examines the training dynamics of the analytic KL variant on science QA. First, it exhibits lower top-$k$ agreement with the self-teacher than SDPO across Physics and Biology, suggesting that the reward signal prevents the policy from collapsing onto the self-teacher's high-probability tokens. Second, it maintains higher entropy than GRPO, with greater levels than SDPO in Physics and comparable levels in Biology, suggesting that reward-driven optimization keeps exploring beyond the teacher's mode.

\subsection{Self-Teacher Design of the Analytic KL Variant}
\label{sec:app_teacher}
\begin{figure}[t]
    \centering
    \includegraphics[trim=1cm 0 1cm 0, clip, width=0.6\textwidth]{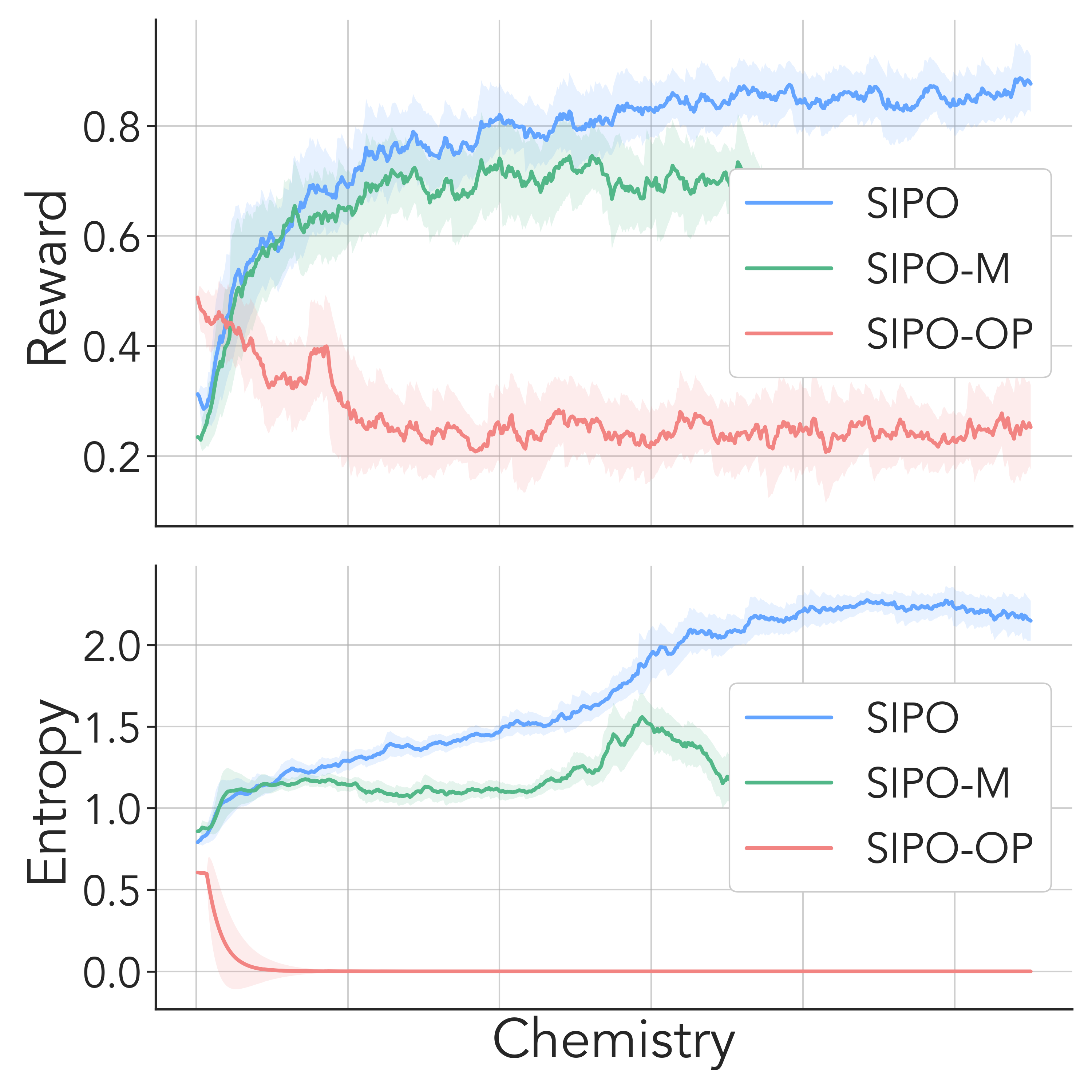}
    \caption{\textbf{Reward and entropy of the analytic KL variant of \ours{} and two ablations on Chemistry.} \ours{}-M conditions the self-teacher on an incorrect instead of a successful rollout, and \ours{}-OP uses the actor's current weights as the teacher. \ours{}-M plateaus at a lower reward, and \ours{}-OP suffers from entropy collapse and fails to learn.}
    \label{fig:teacher}
\end{figure}

We further ablate two design choices of the self-teacher in the analytic KL variant (\Cref{sec:app_analytic}) on Chemistry with Qwen3-8B (\Cref{fig:teacher}). In this variant, the self-teacher is an EMA copy of the policy and is conditioned on a successful in-group rollout as privileged context. \textbf{\ours{}-M} replaces this successful rollout with an incorrect rollout, which the teacher is asked to correct. \textbf{\ours{}-OP} removes the separation between actor and teacher, so that the actor's current weights score the rollouts under the privileged context. Both variants underperform, for different reasons:
(1)~\emph{Successful rollouts are more informative than mistakes as the only privileged context}: \ours{}-M plateaus at a lower reward and its entropy stagnates. A correct rollout gives the teacher a viable reasoning path to anchor its distribution, whereas a mistake mainly signals what not to do and offers limited guidance on its own, so the resulting supervision is weaker and the policy gradient cannot compensate for it. This does not contradict the use of mistakes in the final method, where an incorrect answer is never the only context but serves as the negative context contrasted with the reference answer, so it only needs to indicate which tokens are specific to the mistake.
(2)~\emph{A separate teacher is necessary}: \ours{}-OP fails catastrophically, with the reward stuck at initialization and the entropy soon collapsing to zero, indicating nearly deterministic generation. When the actor and teacher share weights, the divergence term has a trivial minimum at the actor's own distribution and provides no useful gradient, so training reduces to optimizing over low-entropy rollouts and the policy locks onto its initial mode. The final method therefore also keeps the teacher separate from the actor, as a periodic snapshot of the policy synchronized every 10 training steps (\Cref{sec:app_exp}). Overall, effective self-distillation requires both informative privileged context and a teacher that is structurally separate from the actor.

\section{Qualitative Examples}
\label{sec:app_qualitative}

In this section, we present qualitative examples of the self-teacher from an earlier version of \ours{}, in which the teacher is conditioned on a successful in-group rollout (or, in \Cref{fig:qualitative_example_4}, an incorrect one) and generates a response. The final method uses a similar prompt structure (\Cref{fig:prompt}), except that the privileged information is placed before the question and the teacher only re-scores the student's response instead of generating a new one. \Cref{fig:qualitative_example_1,fig:qualitative_example_2,fig:qualitative_example_3} illustrate how the self-teacher leverages a correct solution as privileged information across Chemistry, Physics, and Biology, and \Cref{fig:qualitative_example_4} shows how it analyzes an incorrect reference attempt to deduce the correct final answer on Materials.

\begin{figure}[h]
\begin{prompt}{Chemistry qualitative example}
\texttt{<|im\_start|>user}

Given a question and four options, please select the right answer. Output only the letter corresponding to the correct option (A, B, C, or D), enclosed within <answer>...</answer> tags. For example, if the answer is A, output exactly: <answer>A</answer>

\texttt{ }

Question: How many rotatable bonds are there in the molecule with the canonical SMILES representation 'CC(C(C(=O)NC(C)C(=O)NCC(=O)N1CCCC1C(=O)NC(CCC(=O)N)... \texttt{<omitted>} 

A: 30

B: 26

C: 39

D: 25

\texttt{ }

\texttt{Correct solution:}

To address the task of determining the \textbf{correct number of rotatable bonds} in the molecule described by the SMILES string \texttt{'CC(C(C(=O)...C2CCCN2)O'}: We begin with a chain of structural and functional plasticity... \texttt{<omitted>} ...After reconsulting with the truths as stabilizing norms, the most accurate and verified rotatable bond number is: <answer>A</answer>

\texttt{ }

\texttt{Correctly solve the original question.<|im\_end|>}

\texttt{<|im\_start|>assistant}

\texttt{<|thought|>}
The SMILES string represents a complex peptide-like chain. I must count the single bonds not in a ring, bound to non-terminal heavy atoms, excluding C-N bonds with partial double-bond character... \texttt{<omitted>} 
\texttt{<|thought|>}

To analyze the \textbf{number of rotatable bonds}, we assess the flexible segments versus the rigid polyamide and amide groups. \texttt{<omitted>} Based on formal rules, Option A represents the verified count for this specific molecular topology.

<answer>A</answer>\texttt{<|im\_end|>}
\end{prompt}
\caption{\textbf{Qualitative example of the earlier self-teacher on Chemistry.}}
\label{fig:qualitative_example_1}
\end{figure}

\clearpage

\begin{figure}[h]
\begin{prompt}{Physics qualitative example}
\texttt{<|im\_start|>user}

Given a question and four options, please select the right answer. Output only the letter corresponding to the correct option (A, B, C, or D), enclosed within <answer>...</answer> tags. For example, if the answer is A, output exactly: <answer>A</answer>

\texttt{ }

Question: What type of energy does nuclear decay produce?

A: Nuclear decay produces light and magnetic waves.

B: Nuclear decay produces heat and sound.

C: Nuclear decay produces chemical energy and radiation.

D: Nuclear decay produces energy and gamma rays.

\texttt{ }

\texttt{Correct solution:}

The question of \textbf{energy produced by nuclear decay} is a core component of scientific research... \texttt{<omitted>} ...Option D is the only one that remains consistent with the universal statement of physical reality. \texttt{<answer>D</answer>}

\texttt{ }

\texttt{Correctly solve the original question.<|im\_end|>}

\texttt{<|im\_start|>assistant}

\texttt{<|thought|>}
Nuclear decay involves the transformation of an unstable atomic nucleus. This process releases energy to reach a more stable state. Common emissions include alpha particles, beta particles, and electromagnetic radiation in the form of gamma rays. \texttt{<omitted>} 
\texttt{<|thought|>}

The only scientifically confirmed and universally accepted answer that maintains a complete standard of truth regarding the products of nuclear decay is:

<answer>D</answer>\texttt{<|im\_end|>}
\end{prompt}
\caption{\textbf{Qualitative example of the earlier self-teacher on Physics.}}
\label{fig:qualitative_example_2}
\end{figure}

\clearpage

\begin{figure}[h]
\begin{prompt}{Biology qualitative example}
\texttt{<|im\_start|>user}

Given a question and four options, please select the right answer. Output only the letter corresponding to the correct option (A, B, C, or D), enclosed within <answer>...</answer> tags. For example, if the answer is A, output exactly: <answer>A</answer>

\texttt{ }

What is the relationship between protein ENSP00000258456 and protein ENSP00000336630?

A: The relation is catalysis.

B: The relation is inhibition.

C: The relation is expression. 

D: The relation is ptmod.

\texttt{ }

\texttt{Correct solution:}

To ensure the \textbf{correct, verified, and finalized resolution}, we use the intrinsic signal of Option A, which is appropriate for the binding relationship between TNF and TNFR1... \texttt{<omitted>} ...Therefore, the correct solution is definitively Option A. \texttt{<answer>A</answer>}

\texttt{ }

\texttt{Correctly solve the original question.<|im\_end|>}

\texttt{<|im\_start|>assistant}

\texttt{<|thought|>}
ENSP00000258456 (TNF) and ENSP00000336630 (TNFRSF1A/TNFR1) are key components of the tumor necrosis factor signaling pathway. TNF acts as a ligand that binds to its receptor. Looking at the functional interaction types in protein databases... \texttt{<omitted>} 
\texttt{<|thought|>}

Based on the established functional and biochemical data regarding the interfaces and binding stages of these specific protein sequences, the relationship is identified as catalysis within the signaling cascade.

<answer>A</answer>\texttt{<|im\_end|>}
\end{prompt}
\caption{\textbf{Qualitative example of the earlier self-teacher on Biology.}}
\label{fig:qualitative_example_3}
\end{figure}

\clearpage

\begin{figure}[t]
\begin{prompt}{Materials qualitative example}
\texttt{<|im\_start|>user}

Given a question and four options, please select the right answer. Output only the letter corresponding to the correct option (A, B, C, or D), enclosed within <answer>...</answer> tags. For example, if the answer is A, output exactly: <answer>A</answer>

\texttt{ }

Based on the given data, which of the following materials is predicted to be the most stable perovskite material? Consider the density and the first ionization potential...

Given data:

Material Composition: Sr6Ca2Fe4Ni4O24, Density\_AB\_avg: 5.33975, First Ionization Potential (V)\_AB\_avg: 6.776

Material Composition: Sr6Ca2Fe6Co2O24, Density\_AB\_avg: 5.2115, First Ionization Potential (V)\_AB\_avg: 6.8335

Material Composition: Sr6Ca2Fe6Ni2O24, Density\_AB\_avg: 5.2115, First Ionization Potential (V)\_AB\_avg: 6.805375

Material Composition: Sr6Ca2Fe7Co1O24, Density\_AB\_avg: 5.147375, First Ionization Potential (V)\_AB\_avg: 6.834125

A: Sr6Ca2Fe7Co1O24

B: Sr6Ca2Fe4Ni4O24

C: Sr6Ca2Fe6Co2O24

D: Sr6Ca2Fe6Ni2O24

\texttt{ }

\texttt{Incorrect reference attempt:}

To determine the most stable perovskite... \textbf{highest FIP is in Material A}, so it’s the most stable. \texttt{<omitted>} ...Final Answer: \boxed{A}

\texttt{ }

\texttt{Correctly solve the original question.<|im\_end|>}

\texttt{<|im\_start|>assistant}

\texttt{<|thought|>}
The previous attempt focused solely on the highest FIP (6.834 for Sr6Ca2Fe7Co1O24). However, Sr6Ca2Fe4Ni4O24 (Material B) has the highest density (5.33975) despite a lower FIP (6.776). In many perovskite stability models, density is a primary indicator of structural compactness and thermodynamic stability... \texttt{<omitted>} 
\texttt{<|thought|>}

To determine the most stable material, we must balance the first ionization potential with the material density. While Material A (Sr6Ca2Fe7Co1O24) has a slightly higher FIP, Material B (Sr6Ca2Fe4Ni4O24) exhibits a significantly higher density, suggesting superior structural stability in this specific dataset context.

<answer>B</answer>\texttt{<|im\_end|>}
\end{prompt}
\caption{\textbf{Qualitative example of the earlier self-teacher on Materials.} The self-teacher is given a mistake instead of a correct demonstration as privileged information.}
\label{fig:qualitative_example_4}
\end{figure}

\clearpage


\end{document}